\pdfoutput=1

\documentclass[11pt]{article}

\usepackage[]{acl}

\usepackage{times}
\usepackage{latexsym}

\usepackage[T1]{fontenc}

\usepackage[utf8]{inputenc}

\usepackage{microtype}

\usepackage{amsmath}
\usepackage{mathtools}
\usepackage{enumitem}
\usepackage{booktabs}
\usepackage{graphicx}
\usepackage{color}
\usepackage{comment}
\usepackage{multirow}
\usepackage{multicol}
\usepackage{siunitx}
\usepackage{subcaption}
\usepackage{bm}
\usepackage{soul}
\usepackage{listings}
\newtagform{brackets}{[}{]}
\usetagform{brackets}

\title{Construction Repetition Reduces Information Rate in Dialogue}

\author{Mario Giulianelli\\
  Institute for Logic, Language and Computation \\
  University of Amsterdam \\
  \texttt{m.giulianelli@uva.nl} \\\AND
  Arabella Sinclair\ \ \ \ \ \ \ \ \ \ \ \ \ \ \\
  Department of Computing Science\ \ \ \ \ \ \ \ \ \ \ \ \ \ \\
  University of Aberdeen\ \ \ \ \ \ \ \ \ \ \ \ \ \ \\
  \texttt{arabella.sinclair@abdn.ac.uk}\ \ \ \ \ \ \ \ \ \ \ \ \ \ \\\And
  \ \ \ \ \ \ Raquel Fern\'{a}ndez\\
  \ \ \ \ \ \ Institute for Logic, Language and Computation \\
  \ \ \ \ \ \ University of Amsterdam \\
  \ \ \ \ \ \ \texttt{raquel.fernandez@uva.nl}
}

\begin{document}
\maketitle

\begin{abstract}
	Speakers repeat constructions frequently in dialogue.
	Due to their peculiar information-theoretic properties, repetitions can be thought of as a strategy for cost-effective communication.
	In this study, we focus on the repetition of lexicalised constructions---i.e., recurring multi-word units---in English open-domain spoken dialogues. We hypothesise that speakers use \textit{construction repetition} to mitigate information rate, leading to an overall decrease in utterance information content over the course of a dialogue.
	We conduct a quantitative analysis, measuring the information content of constructions and that of their containing utterances, estimating information content with an adaptive neural language model. We observe that construction usage lowers the information content of utterances. This \textit{facilitating effect} (i)~increases throughout dialogues, (ii)~is boosted by repetition, (iii)~grows as a function of repetition frequency and density, and (iv)~is stronger for repetitions of referential constructions.
\end{abstract}


\section{Introduction}
\label{sec:introduction}

The repeated use of particular configurations of structures and lexemes, \textit{constructions}, is pervasive in conversational language use~\cite{tomasello2003constructing,goldberg2006constructions}.
Such repetition can be understood as a surface level signal of processes of coordination~\cite{sinclair-fernandez-2021-construction} or `interpersonal synergy' between conversational partners~\cite{fusaroli2014dialog}.
Speakers may use repetitions to successfully maintain common ground with their interlocutors~\cite{BrennanClark1996,PickeringGarrod2004}, because they are primed by their recent linguistic experience \cite{bock1986syntactic}, or to avoid a costly on-the-fly search for alternative phrasings~\cite[see, e.g.,][]{kuiper1995smooth}. At the same time, repetitions are also advantageous for comprehenders.
Repeating a sequence of words positively reshapes expectations for those words, allowing comprehenders to process them more rapidly~\cite[for a review, see][]{bigand2005repetition}. As speakers are known to take into consideration both their own production cost and their addressee's processing effort~\cite{clark1986referring,clark1989contributing,frank2012predicting}, its two-sided processing advantage, as described above, makes construction repetition an efficient, cost-reducing communication strategy.
In this paper, we investigate whether and how these information-theoretic properties of repetitions shape patterns of information rate in open-domain spoken dialogue.

Information theory is the study of the conditions affecting the transmission and processing of information. To the foundations of the field belongs the noisy-channel coding theorem~\cite{shannon1948}, which states that for any given degree of noise in a communication channel, it is possible to communicate discrete signals nearly error-free up to a maximum information rate, the \textit{channel capacity}. If speakers use the communication channel optimally, they might send information at a rate that is always close to the channel capacity. This observation is at the basis of the principle of Entropy Rate Constancy~\cite[ERC;][]{genzel2002entropy}, which predicts that the information rate of speaker's utterances, measured as the utterance conditional entropy (i.e., its in-context \textit{Shannon information content} or \textit{information density}) remains constant throughout discourse.
The ERC predictions have been empirically confirmed for written language production~\cite{genzel2002entropy,genzel2003variation,qian2011topic} but results on dialogue are mixed \cite{vega2009looking,doyle2015audience,doyle-frank-2015-shared,xu2018information,giulianelli2021dialogues}, with some studies suggesting a decreasing information rate over the course of dialogues~\cite{vega2009looking,giulianelli2021analysing}.
We hypothesise that this decreasing trend in dialogue may be associated with construction repetition. We conjecture that speakers use construction repetition as a strategy for information rate mitigation, by padding the more information dense parts of their utterances with progressively less information dense constructions---leading to an overall decrease in information rate over the course of a dialogue. \looseness=-1

We extract occurrences of fully lexicalised constructions (see Table~\ref{tab:construction-examples} for examples) from a corpus of open-domain spoken dialogues and use a Transformer-based neural language model to estimate their contribution to utterance information content.
First, we confirm that constructions indeed exhibit lower information content than other expressions and that information content further decreases when constructions are repeated. Then, we show that the decreasing trend of information content observed \textit{over utterances}---which contradicts the ERC principle---is driven by the increasing mitigating effect of construction repetition, measured as a construction's (increasingly) negative contribution to the information content of its containing utterance, what we call its \textit{facilitating effect}.

In sum, our study provides new empirical evidence that dialogue partners use construction repetition as a strategy for information rate mitigation, which can explain why the rate of information transmission in dialogue, in contrast to the constancy predicted by the theory~\cite{genzel2002entropy}, is often found to decrease.
Our findings inform the development of better dialogue models. They indicate, as suggested in related work \cite[e.g.,][]{xi2021taming}, that while avoiding \textit{degenerate} repetitions in utterance generation \cite{li2016deep,welleck2019neural} is an appropriate strategy, dialogue systems should not suppress \textit{human-like} patterns of repetition as these make automatic systems be perceived as more natural and more effective in conversational settings.


\section{Background}
\label{sec:background}

\subsection{Constructions}
\label{sec:background-constructions}
This work focuses on \textit{constructions}, seen as particular configurations of structures and lexemes in usage-based accounts of natural language \cite{tomasello2003constructing,bybee2006usage,bybee2010language,goldberg2006constructions}.
According to these accounts, models of language processing must consider not only individual lexical elements according to their syntactic roles but also more complex form-function units, which can break regular phrasal structures---e.g., \textit{`I~know I'}, \textit{`something out of'}.
We further focus on fully lexicalised constructions (sometimes called \textit{formulaic expressions}, or \textit{multi-word expressions}).
Commonly studied types of constructions are idioms (\textit{`break the ice'}), collocations (\textit{`pay attention to'}), phrasal verbs (\textit{`make up'}), and lexical bundles (\textit{`a lot of the'}). In §\ref{sec:extraction}, we explain how the notion of lexicalised construction is operationalised in the current study; Table~\ref{tab:construction-examples} shows some examples.

\begin{table}[t]
\centering
\small
\resizebox{\columnwidth}{!}{
\begin{tabular}{@{}p{0.151\textwidth} p{0.155\textwidth} p{0.134\textwidth}@{}}
\toprule
\textbf{SPXV}    	       & \textbf{SAXQ}                            &  \textbf{S9YG} \\ \midrule
want to be with him    &  \textit{it on the television}       &  I bet you can \\
\textit{shit like that}    &  \textit{for a family}  				  &  yeah I used to \\
I can be  						&  think that's a  							  &  \textit{go to bed}  \\
to see her  				  &  \textit{the orient express}  	   &  and I love \\
and she just  				&  one thing that  							&  \textit{the window and} \\
I quite like  					&  \textit{one of my favourites}  &  and I think it's \\
you don't like  		    &  \textit{on the television}  		  &  yeah I think so \\
and you're like 		   &  yes yeah I  								 &  \textit{the same people} \\
going to go  				&  erm I think  							  &  is she in \\
you're going to  		 &  a really good  							 &  \textit{lock the door} \\ \bottomrule
\end{tabular}
}
\caption{Top 10 constructions from three dialogues of the Spoken BNC \cite{love2017spoken}, sorted according to the PMI between a construction and its dialogue (§\ref{sec:stat-model}). Referential constructions in italics (§\ref{sec:extraction}). Headers correspond to the dialogues' IDs in the corpus.}
\label{tab:construction-examples}
\end{table}

A common property of constructions is their frequent occurrence in natural language. As such, they possess what, in usage-based accounts, is sometimes referred to as `processing advantage' \cite{conklin2012processing,carrol2020all}. Evidence for the processing advantage of construction usage has been found in reading \cite{arnon2010more,tremblay2011processing}, naming latency \cite{bannard2008stored,janssen2012phrase}, eye-tracking \cite{underwood2004eyes,siyanova2011seeing}, and electrophysiology \cite{tremblay2010holistic,siyanova2017representation}.
In this paper, we model this processing advantage as reduced information content and show that it can mitigate information rate throughout entire dialogues.

\subsection{Information Content, Surprisal, and Processing Effort}
\label{sec:background-surprisal}

Estimates of information content have been shown to be good predictors of processing effort in perception \cite{jelinek1975design,clayards2008perception}, reading \cite{keller2004entropy,demberg2008data,levy2009eye}, and sentence interpretation \cite{levy2008noisy,gibson2013rational}.
In these studies, information content is typically referred to as \textit{surprisal}, taken as a measure of how unpredictable, unlikely, or surprising a linguistic signal is in its context.
As speakers take into consideration their addressee's processing effort \cite{clark1986referring,clark1989contributing}, their linguistic choices can often be explained as strategies to manage the fluctuations of information content over time. Surprisal-based accounts have indeed been successful at explaining various aspects of language production: speakers tend to reduce the duration of less surprising sounds \cite{aylett2004smooth,aylett2006language,bell2003effects,demberg2012syntactic}; they are more likely to drop sentential material within less surprising scenarios \cite{jaeger2007speakers,frank2008speaking,jaeger2010redundancy};
they tend to overlap at low-surprisal dialogue turn transitions \cite{dethlefs2016information}; and they produce sentences at a constant information rate in texts \cite{genzel2002entropy,qian2011topic,giulianelli2021analysing}.

To measure information content we use GPT-2 \cite{radford2019language}, a neural language model.
We thereby follow the established approach~\cite[e.g.,][]{genzel2002entropy,keller2004entropy,xu2018information} of using language models to estimate information content.
Neural models' estimates in particular have been shown to be good predictors of processing effort, measured as reading time, gaze duration, and N400 response \cite{monsalve2012lexical,goodkind-2018-predictive,merkx2021human,schrimpf2021neural}.
We further implement a simple neural adaptation mechanism, performing continuous gradient updates based on utterance prediction error; this not only leads to a more psychologically plausible model but also to the estimation of more human-like expectations \cite{van2018neural}.\looseness=-1


\section{Data}
\label{sec:data}
We conduct our study on the Spoken British National Corpus\footnote{\url{http://www.natcorp.ox.ac.uk}.}~\cite{love2017spoken}, a dataset of transcribed open-domain spoken dialogues containing 1,251 contemporary British English conversations, collected in a range of real-life contexts. We focus on the 622 dialogues that feature only two speakers, and randomly split them into a 70\% finetuning set (to be used as described in §\ref{sec:method}) and a 30\% analysis set (used in our experiments, as described in §\ref{sec:preliminary} and~§\ref{sec:results}). Table~\ref{tab:bnc-length} shows some statistics of the dialogues used in this study.

\begin{table}[h]
\centering
\resizebox{\columnwidth}{!}{
\begin{tabular}{@{}lcccc@{}}
\toprule
                                    & \textbf{Mean $\pm$ Sd} & \textbf{Median} & \textbf{Min} & \textbf{Max} \\ \midrule
\textbf{Dialogue length (\# utterances)} & 736 $\pm$ 599           & 541.5             & 67          & 4859         \\[2pt]
\textbf{Dialogue length (\# words)} & 7753 $\pm$ 5596         & 6102            & 819         & 39575        \\[2pt]
\textbf{Utterance length (\# words)}     & 11 $\pm$ 15             & 6               & 1            & 982    \\ \bottomrule
\end{tabular}
}
\caption{
Two-speaker dialogue statistics, Spoken BNC.}
\label{tab:bnc-length}
\end{table}

\subsection{Extracting Repeated Constructions}
\label{sec:extraction}
We define constructions as multi-word sequences repeated within a dialogue.
To extract constructions from each dialogue, we use the sequential pattern mining method proposed by \citet{duplessis2017utterance,duplessis2017automatic,duplessis2021towards}, which treats the extraction task as an instance of the longest common subsequence problem \cite{hirschberg1977algorithms,bergroth2000survey}.\footnote{Their code is freely available at \url{https://github.com/GuillaumeDD/dialign}.} We modify it to not discard multiple repetitions of a construction that occur in the same utterance.
We focus on constructions of at least three tokens, uttered at least three times in a dialogue by any of the dialogue participants.
Repeated sequences that mostly appear as a sub-part of a larger construction are discarded.\footnote{We discard constructions that appear less than twice outside of a larger repeated construction in a given dialogue (e.g., \textit{`think of it'} vs.\ \textit{`think of it like'}).}
We also exclude sequences containing punctuation marks or which consist of more than 50\% filled pauses (e.g., \textit{`mm'}, \textit{`erm'}).\footnote{The full list of filled pauses can be found in Appendix~\ref{sec:app-extraction}.}

Applying the described extraction procedure to the 187 dialogues in the analysis split of the Spoken BNC yields a total of 5,893 unique constructions and 60,494 occurrences. Further statistics of the extracted constructions are presented in Table~\ref{tab:construction-stats}, and Table~\ref{tab:construction-examples} shows 10 example constructions extracted from three dialogues.
For analysis purposes, we distinguish between referential and non-referential constructions. We label a construction as \textit{referential} if it includes nouns, unless the nouns are highly generic.\footnote{We define a limited specific vocabulary of generic nouns (e.g., \emph{`thing'}, \emph{`fact'}, \emph{'time'}); full vocabulary in Appendix~\ref{sec:app-extraction}.} Referential constructions are mostly topic-determined; examples are \emph{`playing table tennis'}, \emph{`a woolly jumper'}, \emph{`a room with a view'}. The remaining constructions are labelled as \textit{non-referential}. These mainly include topic-independent expressions and conversational markers, such \emph{`a lot of'}, \emph{`I don't know'}, and \emph{`yes of course'}. Our dataset consists of 5,291 referential and 55,203 non-referential construction occurrences, 1,143 and 4,750 construction forms; see Table~\ref{tab:construction-examples} for further examples.

\begin{table}[h]
\centering
\resizebox{\columnwidth}{!}{
	\begin{tabular}{@{}lccc@{}}
	\toprule
	                                                  & \textbf{Mean $\pm$ Sd} & \textbf{Median} & \textbf{Max} \\ \midrule
	\textbf{Construction Length}                    & 3.27 $\pm$ 0.58             & 3                       & 7         \\[2pt]
	\textbf{Construction Frequency}             & 4.29 $\pm$ 3.04              & 3                   & 70        \\[2pt]
	\textbf{Constructions per Dialogue}       & 325.34 $\pm$ 458.64              & 149            & 2817        \\[2pt]
	\ \ \ \ \small \textbf{\textit{Referential}}         &  \small 30.96 $\pm$ 39.75              &  \small 19                         &  \small 346        \\[2pt]
	\ \ \ \ \small \textbf{\textit{Non-Referential}} &  \small 296.88 $\pm$ 424.17              &  \small 134.5                         &  \small 2530        \\[2pt]
	\textbf{Utterance Length}                  & 31.19 $\pm$ 36.19                  & 21                     & 959    \\ \bottomrule
	\end{tabular}
}
\caption{Construction statistics for our analysis split of the Spoken BNC. \textit{Constr.\ Frequency}: occurrences of a given construction in a dialogue. \textit{Constr.\ per Dialogue}: occurrences of all constructions in a dialogue. \textit{Utterance Length}: number of words in utterances containing a construction. The minimum is always 3 by design (§\ref{sec:extraction}). The difference between referential and non-referential is only significant for \textit{Constr.\ per Dialogue}.}
\label{tab:construction-stats}
\end{table}


\section{Experimental Setup}
\label{sec:method}
In this section, we define our information-theoretic measures and present the adaptive language model used to produce information content estimates.\footnote{Code and statistical analysis are available at \url{https://github.com/dmg-illc/uid-dialogue}.}

\subsection{Information Content Measures}
\label{sec:estimates}
The \textit{information content} of a word choice $w_i$ is the negative logarithm of the corresponding word probability, conditioned on the utterance context~$u_{:w_i}$ (i.e., the words that precede $w_i$ in utterance~$u$) and on the local dialogue context $l$:
\begin{align}
    H(w_i|u_{:w_i},l) = - \log_2 P(w_i | u_{:w_i},l)
    \label{eq:word-surprisal}
\end{align}
We define the local dialogue context~$l$ as the 50 tokens that precede the first word in the utterance.\footnote{Building on prior work \cite{reitter2006computational} that uses a window of 15 seconds of spoken dialogue as the locus of local repetition effects, we compute the average speech rate in the Spoken BNC (3.16 tokens/second) and multiply it by 15; we then round up the result (47.4) to 50 tokens.}
We use tokens as a unit of context size, rather than utterances, since they more closely correspond to the temporal units used in previous work \cite[e.g.,][]{reitter2006computational}, and since the length of utterances can vary significantly (see Table~\ref{tab:bnc-length}).
To measure the information content of a construction~$c$, we average over word-level information content values:
\begin{align}
    H(c;u_{:c},l) = \frac{1}{|c|} \sum_{w_i \in c} H(w_i|u_{:c},l)
    \label{eq:construction-surprisal}
\end{align}
We use the same averaging strategy to compute the information content of entire utterances, following prior work~\cite[e.g.,][]{genzel2002entropy,xu2018information}:
\begin{align}
	H(u;l) = \frac{1}{|u|} \sum_{w_i \in u} H(w_i|u_{:w_i},l)
	\label{eq:utterance-surprisal}
\end{align}

The above information content estimates target constructions and entire utterances but they do not qualify the relationship between the two.
We also measure the information content change (increase or reduction in information rate) contributed by a construction $c$ to its containing utterance, which we call the \textit{facilitating effect} of a construction.
Facilitating effect is defined as the logarithm of the ratio between the information content of a construction and that of its utterance context:
\begin{align}
 {\it FE}(c;u,l) = \log_2 \frac{\frac{1}{|u|-|c|} \sum_{c \not\ni w_j \in u} H(w_j|u_{:w_i},l)}{\frac{1}{|c|} \sum_{w_i \in c} H(w_i|u_{:c},l)}
\label{eq:facilitating-effect}
\end{align}
By definition, this quantity is positive when the construction has lower information content than its context, and negative when it has higher information content.
When the utterance consists of a single construction, facilitating effect is set to 0.

We can expect the values produced by our information content and facilitating effect measurements (Eq.\ \ref{eq:construction-surprisal} and \ref{eq:facilitating-effect}, respectively) to correlate: it is more likely for a construction to have a (positive) facilitating effect if its information content is low. When a construction's information content is high, the information content of its utterance context must be even greater for facilitating effect to occur. Nevertheless, perfect correlation does not follow a priori from the definition of the two measures; we will show this empirically in §\ref{sec:ic-vs-fe}.

\subsection{Language Model}
\label{sec:modelling}

To estimate the per-word conditional probabilities that are necessary to compute information content (Eq.\ \ref{eq:word-surprisal}), we use an adaptive language model. The model is conditioned on local contextual cues via an attention mechanism \cite{vaswani2017attention} and it learns continually \cite[see, e.g.,][]{krause2018dynamic} from exposure to the global dialogue context.
We use GPT-2 \cite{radford2019language}, a pre-trained autoregressive Transformer language model.
We rely on HuggingFace's implementation of GPT-2 with default tokenizers and parameters \cite{wolf2020transformers} and finetune the pre-trained model on a 70\% training split of the Spoken BNC to adapt it to the idiosyncrasies of spoken dialogic data.\footnote{More details on finetuning can be found in Appendix~\ref{sec:app-finetuning}.}
We refer to this finetuned version as the \textit{frozen} model.
We use an attention window of length $|u_{:w_i}| + 50$, i.e., the sum of the utterance length up to word $w_i$ and the size of the local dialogue context.

As a continual learning mechanism, we use back-propagation on the cross-entropy next word prediction error, a simple yet effective adaptation approach motivated in §\ref{sec:background-surprisal}.
Following \citet{van2018neural}, when estimating information content for a dialogue, we begin by processing the first utterance using the frozen language model and then gradually update the model parameters after each turn.
For these updates to have the desired effect, the learning rate should be appropriately tuned. It should be sufficiently high for the language model to adapt during a single dialogue, yet an excessively high learning rate can cause the language model to lose its ability to generalise across dialogues.
To find the appropriate rate, we randomly select 18 dialogues from the analysis split of the Spoken BNC\footnote{This amounts to ca.~10\% of the analysis split. We use the analysis split because there is no risk of ``overfitting'' with respect to our main analyses.} and run an 18-fold cross-validation for a set of six candidate learning rates: \num{1e-5}, \num{1e-4}, $\ldots$,~1. We finetune the model on each dialogue using one of these learning rates and compute perplexity reduction (i)~on the dialogue itself (\textit{adaptation}) as well as (ii)~on the remaining 17 dialogues (\textit{generalisation}).
We select the learning rate yielding the best adaptation over cross-validation folds (\num{1e-3}), while still improving the model's generalisation ability. See Appendix~\ref{sec:app-lr} for further details.
\looseness=-1


\section{Preliminary Experiments}
\label{sec:preliminary}
In this section, we present preliminary experiments on the information content of utterances and constructions, which set the stage for our analysis of the facilitating effect of construction repetition.

\subsection{Utterance Information Content}
\label{sec:preliminary-utterance}
\looseness=-1
Our experiments are motivated by the mixed results on the dynamics of information rate in dialogue discussed in §\ref{sec:introduction}. We thus begin by testing if the Entropy Rate Constancy (ERC) principle holds in the Spoken BNC, i.e., whether utterance information content remains stable over the course of a dialogue.
Following a procedure established in prior work~\cite{xu2018information}, we fit a linear mixed effect model with the logarithm of utterance position and construction length as fixed effects (we will refer to their coefficients as $\beta$), and include multi-level random effects grouped by dialogue. \textit{For the ERC principle to hold, the position of an utterance within a dialogue should have no effect on its information content.}
Instead, we find that utterance information content decreases significantly over time ($\beta\!=\!-0.119, p\!<\!0.005$, \textit{95\% c.i.}\ $-0.130\!:\!-0.108$), in line with previous negative results on open-domain and task-oriented dialogue~\cite{vega2009looking,giulianelli2021analysing}.
The strongest drop occurs in the first ten dialogue utterances ($\beta\!=\!-0.886, p\!<\!0.005$, \textit{95\% c.i.}\ $-0.954\!:\!-0.818$) but the decrease is still significant for later utterances ($\beta\!=\!-0.043, p\!<\!0.005$, \textit{95\% c.i.}\ $-0.054\!:\!-0.032$).

\subsection{Construction Information Content}
\label{sec:preliminary-construction}
Our hypothesis that construction repetition progressively reduces the information rate of utterances is motivated by the fact that constructions are known to have a processing advantage (see §\ref{sec:introduction} and §\ref{sec:background-constructions}).
This property makes them an efficient production strategy, i.e., one that reduces the speaker's and addressee's collaborative effort. Before investigating if the hypothesised information rate mitigation strategy is at play, we test whether our information theoretic measures and the language model used to generate them are able to capture processing advantage: \textit{we expect our framework to yield lower information content estimates (Eq.\ \ref{eq:construction-surprisal}) for constructions than for other word sequences.}
Indeed, the information content of constructions is significantly lower than that of non-construction sequences ($t\!=\!-168.82, p\!<\!0.005$, \textit{95\% c.i.}\ $-2.033\!:\!-1.987$).\footnote{We extract all 3- to 7-grams from our analysis split of the Spoken BNC, excluding all \textit{n}-grams that are equal to extracted constructions. We then sample, for each length $n$ from 3 to 7, $s_n$ non-construction sequence occurrences---where $s_n$ is the number of occurrences of $n$-tokens-long extracted constructions.\label{ftn:non-constructions}. The length distributions should match because length has an effect on \textit{S} and \textit{FE} (see §\ref{sec:construction-types}).}
Constructions' information content is on average 2 bits lower than that of non-constructions.
We conclude that our estimates of information content are a sensible model of the processing advantage of constructions.

\subsection{Stable Rate of Construction Usage}
\label{sec:stable-usage}
In experiment §\ref{sec:preliminary-construction}, we confirmed that constructions have lower information content than other utterance material. A simple strategy to decrease utterance information content over dialogues (we do observe this decrease in the Spoken BNC, as described in §\ref{sec:preliminary-utterance}) could then simply be to increase the rate of construction usage.
To test if this strategy is at play, we fit a linear mixed effect model with utterance position as the predictor and the proportion of construction tokens in an utterance as the response variable.
Over the course of a dialogue, the increase in the proportion of an utterance's tokens which belong to a construction is negligible ($\beta\!=\!0.004, p\!<\!0.05$, \textit{95\% c.i.}\ $0.001\!:\!0.008$).
Speakers produce constructions at a stable rate (see also Figure~\ref{fig:proportions} in Appendix~\ref{sec:app-extraction}), indicating that an alternative strategy for information rate reduction is at work.\looseness-1

\begin{table*}
	\centering
	\small
	\resizebox{\textwidth}{!}{
		\begin{tabular}{cccclccc}
			\toprule
			\textbf{Speaker} & \textbf{RI} & \textbf{RI~Utt} & \textbf{Dist} & \textbf{Turn}  & $\boldsymbol{H(u)}$ & $\boldsymbol{H(c)}$ & $\boldsymbol{FE(c;u)}$ \\
			\midrule
			A  &  0  &  0  &  -  &  Drink? that was what he did yeah just just to just to know that & 5.99 & 4.73 & 0.40 \\
			& & & &	 \ \ I he \textbf{might not be} a complete twat but just a fyi  & & & \\ \midrule
			B & 1 & 0 & 1586 &  Especially for my birthday mind you I \textbf{might not be} here for & 5.04 & 4.01 & 0.53 \\
			& 2 & 1 & 14 &  \ \ mine and I went what do you mean you \textbf{might not be} here?  & & 2.70 & 0.90\\ \bottomrule
		\end{tabular}
	}
	\caption{
		Repetition chain for the construction \textit{`might not be'} in dialogue SXWH of the Spoken BNC, annotated with repetition index (RI), RI in utterance (RI~Utt), and distance from previous mention (Dist; in tokens). $H(u)$ is the utterance information content, $H(c)$ and $FE(c;u)$ are the construction's information content and facilitating effect.
}
\label{tab:chain-len3}
\end{table*}

\subsection{Information Content vs.\ Facilitating Effect}
\label{sec:ic-vs-fe}
The facilitating effect \textit{FE} of a construction is a function of its information content and the information content of its containing utterance~(Eq.~\ref{eq:facilitating-effect}).
To ensure that our estimates of \textit{FE} are not entirely determined by construction information content (cf.\ §\ref{sec:estimates}), we inspect the relation between the two measures empirically, by looking at the values they take in our dataset of constructions.
We find that the Kendall's rank-correlation between \textit{FE} and information content is $-0.623$ ($p\!<\!0.005$): although this is a rather strong negative correlation, the fact that the score is not closer to $-1$ indicates that there are cases where the two values are both either high or low.
We indeed find examples of constructions with high information content \textit{H} and high facilitating effect \textit{FE}:
\vspace{0.4em} \\
\noindent
\resizebox{\columnwidth}{!}{
	\begin{tabular}{l@{\ }l@{}}
		A:        & we'll level that right press p purchase and     \\
		B:        & right \\
		A:        & \textbf{go back to} recommended \textit{\textbf{(}}$\boldsymbol{H\!=\!5.30 \ \ FE\!=\!1.65}$\textit{\textbf{)}}  \ \ \    \vspace{0.8em}  \\
	\end{tabular}
}
as well cases where information content is low and facilitating effect is low or negative:
\vspace{0.4em} \\
\noindent
\resizebox{\columnwidth}{!}{
	\begin{tabular}{l@{\ }l@{}}
		A:        & right let's go and have a drink \\
		B           & yeah    \\
		A:        & \textbf{let's go and have} a drink   \textit{\textbf{(}}$\boldsymbol{H\!=\!2.10 \ \ FE\!=\!-2.21}$\textit{\textbf{)}} \vspace{0.8em}  \\
	\end{tabular}
}
These examples have been selected among occurrences with \textit{H}/\textit{FE} higher or lower than the mean \textit{H}/\textit{FE} $\pm$ sd, respectively $3.62\!\pm\!1.48$ and $0.62\!\pm\!0.73$.
Further analysis shows that this is not only true for individual instances but for entire groups of constructions. In particular, although their information content is overall higher ($t\!=\!13.511, p\!<\!0.005$, \textit{95\% c.i.}\ $0.371\!:\!0.497$), referential constructions also have higher facilitating effect than non-referential ones ($t\!=\!3.115, p\!<\!0.005$, \textit{95\% c.i.}\ $0.016\!:\!0.072$).
We conclude that the two measures capture different aspects of a construction's information rate profile, with facilitating effect being sensitive to both construction and utterance information content.


\begin{figure*}[h]
	\centering
	\begin{subfigure}{.23\textwidth}
		\centering
		\includegraphics[width=\linewidth]{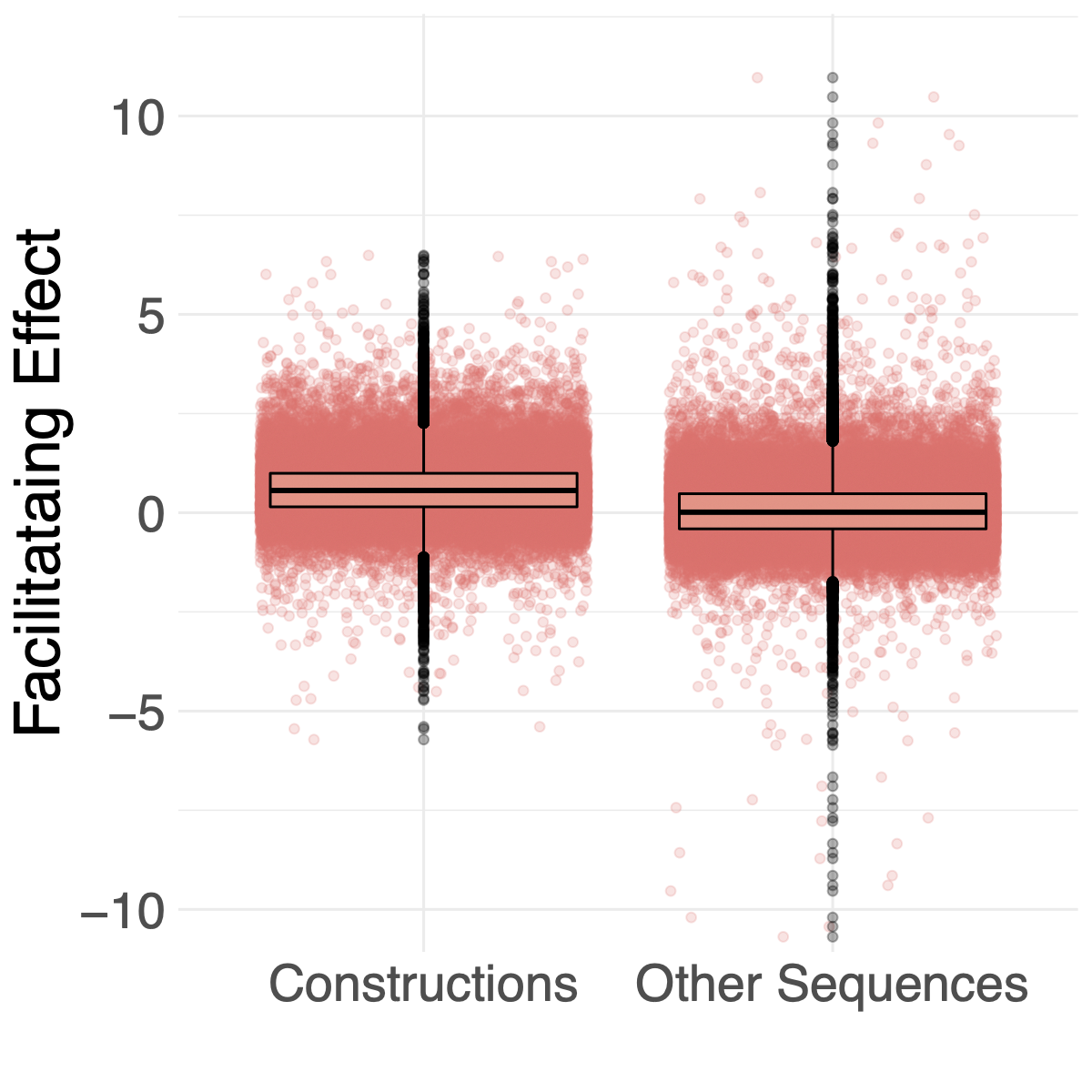}
		\caption{}
		\label{fig:fe-constr-nonconstr}
	\end{subfigure} \hspace{0.5em}
	\begin{subfigure}{.23\textwidth}
		\centering
		\includegraphics[width=\linewidth]{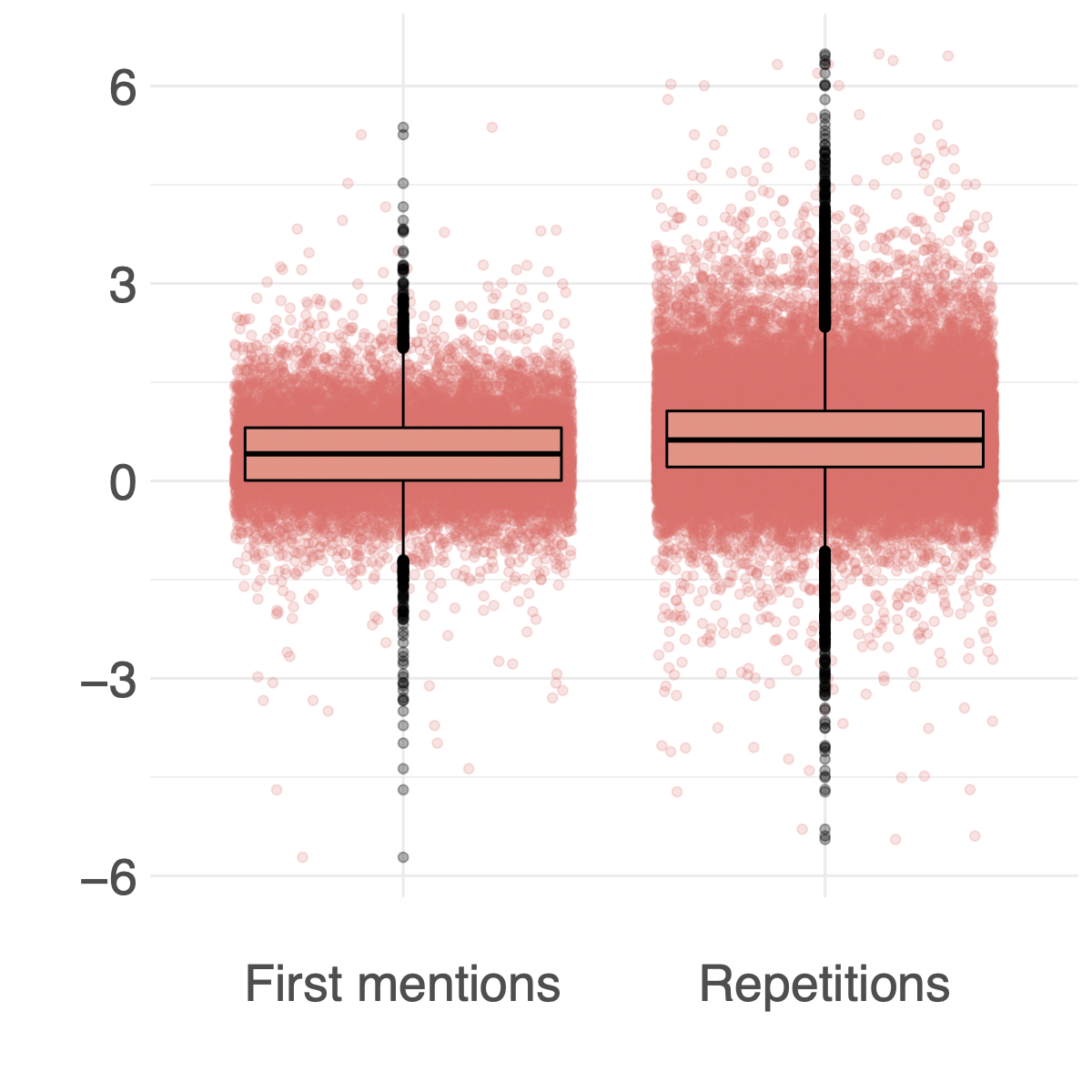}
		\caption{}
		\label{fig:fe-rep-first}
	\end{subfigure} \hspace{0.5em}
	\begin{subfigure}{.23\textwidth}
		\centering
		\includegraphics[width=\linewidth]{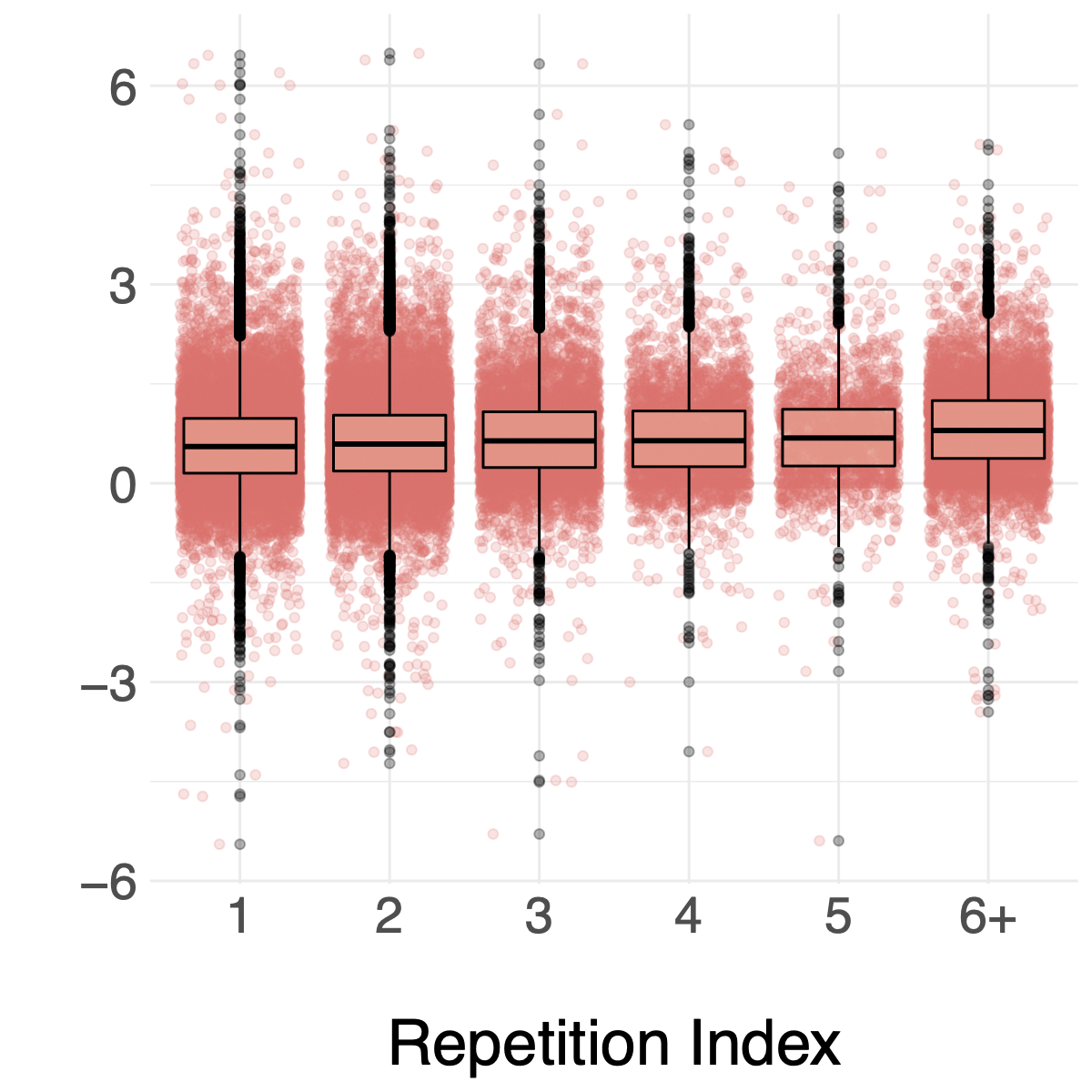}
		\caption{}
		\label{fig:fe-cumul}
	\end{subfigure} \hspace{0.5em}
	\begin{subfigure}{.23\textwidth}
		\centering
		\includegraphics[width=\linewidth]{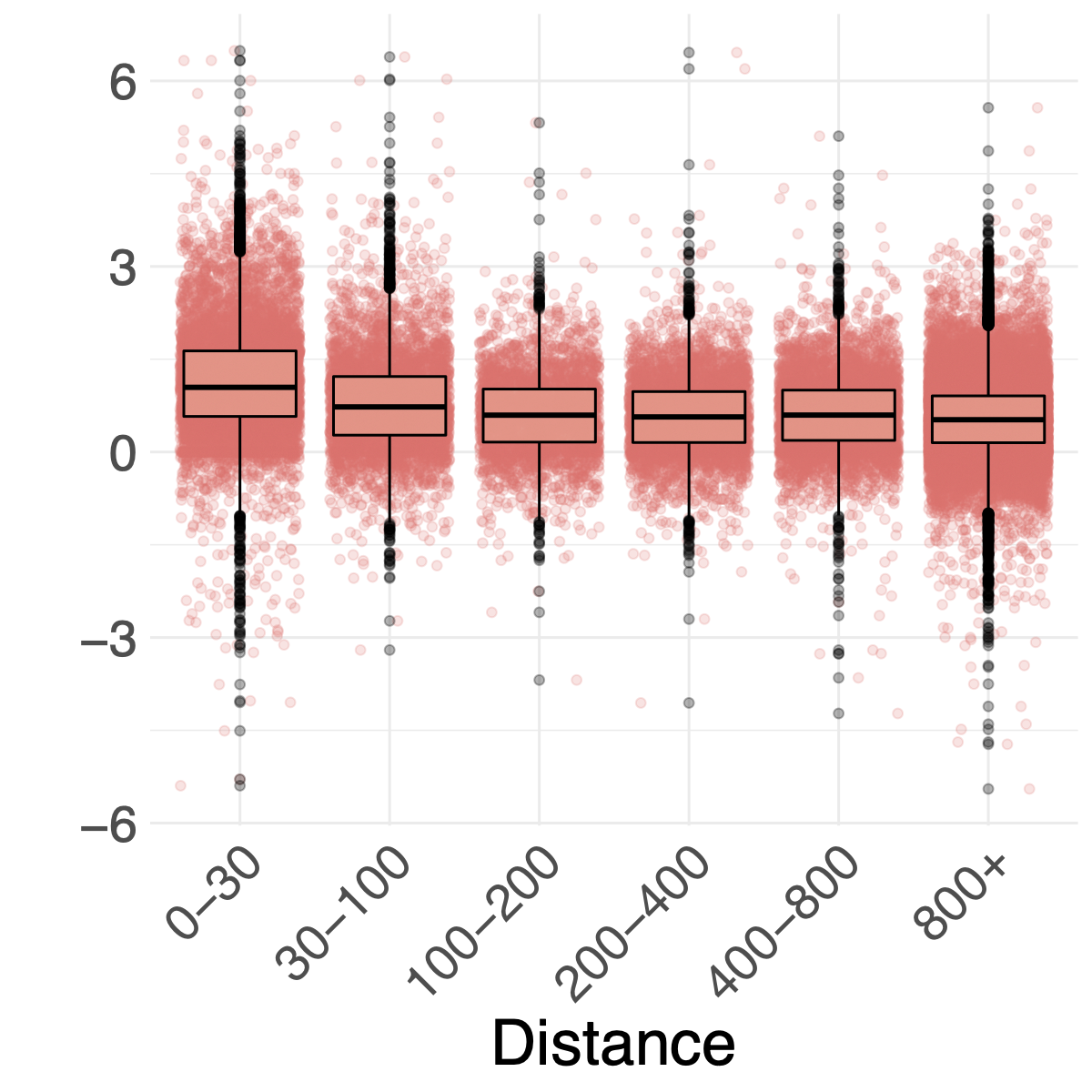}
		\caption{}
		\label{fig:fe-decay}
	\end{subfigure}
	\caption{The facilitating effect (\textit{FE}) of constructions vs.\ non-construction sequences (a) and of first construction mentions vs.\ repetitions (b); as well as \textit{FE} vs.\ repetition index (c) and \textit{FE} vs.\ distance from previous mention (number of words). The first distance bin is the mean length of a turn containing a construction (Table~\ref{tab:construction-stats}).}
	\label{fig:main-plots}
\end{figure*}

\section{The Facilitating Effect of Construction Repetition}
\label{sec:results}
We now test whether constructions have a positive facilitating effect, i.e., whether they reduce the information content of their containing utterances.
We present our main statistical model in §\ref{sec:stat-model}, describe the effects of \textit{FE} predictors specific to unique construction mentions in §\ref{sec:results-mentions}, and analyse differences between types of constructions in §\ref{sec:construction-types}.\looseness-1

\subsection{Method}
\label{sec:stat-model}
\looseness=-1
To understand what shapes a construction's facilitating effect, we collect several of motivated features that can be expected to be informative \textit{FE} predictors. We fit a linear mixed effect (LME) model using (i)~these features as fixed effects, (ii)~\textit{FE} as the response variable, (iii)~and multi-level random effects grouped by dialogue and individual speaker ID.
The first predictor is \textit{utterance position}, i.e., the index of the utterance within the dialogue, which allows us to test if \textit{FE} increases over the course of a dialogue. We then include predictors that distinguish different types of repetition.
Since we expect a construction mention to increase expectation for subsequent occurrences---thus reshaping their information content---we consider its \textit{repetition index}, i.e., how often the construction has been repeated so far in the dialogue.
Expectation is also shaped by intervening material, so we additionally track \textit{distance}, the number of tokens separating a construction mention from the preceding one. As \textit{FE} is the interplay between a construction and its utterance context, it is important to know whether the utterance context contains other mentions of the construction. We use a binary indicator (\textit{previous same utterance}) to single out occurrences whose previous mention is in the same utterance; for these cases, we also count the number of same-utterance previous mentions (\textit{repetition index in utterance}).
To explore whether \textit{FE} varies across types of expressions, we also include a binary feature indicating whether the construction is \textit{referential} or non-referential (§\ref{sec:extraction}).
Finally, we keep track of \textit{construction length}, the number of tokens that constitutes a construction, and \textit{PMI}, the pointwise mutual information between a construction and its dialogue, which is essentially a measure of the construction's frequency in the current dialogue as a function of its overall frequency in the corpus, indicating the construction's degree of interaction-specificity.\footnote{
	The probabilities for the PMI calculation are obtained using maximum likelihood estimation over our analysis split of the Spoken BNC.}

To determine the fixed effects of the final model, we start with all the predictors listed above (the non-binary ones are log-transformed) and perform backward stepwise selection, iteratively removing the predictor with the lowest significance and keeping only those with $p\!<\!0.05$. All predictors make it into our final model, the one which best fits the data according to both the Akaike and the Bayesian Information Criterion.
The full specification of the best model, with model fit statistics as well as fixed and random effect coefficients, are in Appendix~\ref{sec:app-lme}. The next two sections present our main findings; we report fixed effect coefficients~($\beta$), p-values~($p$), and 95\% confidence intervals~(\textit{c.i.}).\looseness-1

\subsection{Construction Mentions}
\label{sec:results-mentions}
Our first observation is that construction usage reduces \textit{utterance} information content. More precisely, we find that \textbf{facilitating effect is higher for constructions than for non-construction sequences} ($t\!=\!118.79, p\!<\!0.005$, \textit{95\% c.i.}\ $0.536\!:\!0.554$).
Constructions have on average 62\% lower information content than their utterance context; the average percentage drops to 7\% for non-construction sequences.\footnote{
These are the same sampled non-construction sequences as in~§\ref{sec:preliminary-construction}. Their average \textit{FE} is $0.07 \pm 0.80$.
}
Figure~\ref{fig:fe-constr-nonconstr} shows the two distributions.
We also observe a positive effect of utterance position on \textit{FE} ($\beta\!=\!0.046, p\!<\!0.005$, \textit{95\% c.i.}\ $0.026\!:\!0.06$); that is, \textbf{the facilitating effect of constructions increases over the course of dialogues}. While the proportion of construction tokens remains stable~(§\ref{sec:stable-usage}), their mitigating contribution to utterance information content increases throughout dialogues---perhaps since speakers are more likely to \textit{repeat} established constructions as the dialogue develops.
We indeed find that \textbf{repeated constructions have stronger facilitating effect}: there is a significant difference between the \textit{FE} of first mentions and repetitions ($t\!=\!-38.904, p\!<\!0.005$, \textit{95\% c.i.}\ $-0.265\!:\!-0.239$), as shown in Figure~\ref{fig:fe-rep-first}. The information content of repetitions is on average 68\% lower than that of their utterance context; for first mentions, it is on average 42\% lower.

Having observed that the mitigating contribution of constructions to utterance information content indeed increases with construction repetition, we now look at how the \textit{FE} of repetitions varies as a function of their distribution across time.
On the one hand, we find that \textbf{facilitating effect is cumulative}: repeating a construction reduces utterance information content more strongly as more mentions of the construction accumulate in the dialogue (Figure~\ref{fig:fe-cumul}). The effect of repetition index (i.e., how often the construction has been repeated so far in the dialogue) is positive on \textit{FE} ($\beta\!=\!0.079, p\!<\!0.005$, \textit{95\% c.i.}\ $0.063\!:\!0.094$).
On the other hand, the distance of a repetition from the previous mention has a negative effect on \textit{FE} ($\beta\!=\!-0.311, p\!<\!0.005$, \textit{95\% c.i.}\ $-0.328\!:\!-0.293$). That is, \textbf{facilitating effect decays as a function of the distance between subsequent mentions}.
As shown in Figure~\ref{fig:fe-decay}, this is a fast decay effect: the most substantial drop occurs for low distance values. The large magnitude of this coefficient indicates that recency is an important factor for constructions to have a strong facilitating effect. Indeed, almost one third (31.8\%) of all repetitions produced by speakers are not more than 200 tokens apart from their previous mention.
Further results showing strong cumulativity effects for self-repetitions within the same utterance can be found in Appendix~\ref{sec:app-same-utterance}.

\subsection{Types of Construction}
\label{sec:construction-types}
In this section, we analyse factors shaping the facilitating effect of construction forms, rather than individual mentions. We focus on the length of a construction and on whether it is referential.

Construction length has a positive effect on \textit{FE} ($\beta\!=\!0.098, p\!<\!0.005$, \textit{95\% c.i.}\ $0.087\!:\!0.119$): \textbf{longer constructions have stronger facilitating effect.} Table~\ref{tab:chain-len3} shows a full repetition chain for a construction of length 3; Table~\ref{tab:chain-len7} (Appendix~\ref{sec:app-extraction}) for one of length 6.
Non-construction sequences display an opposite, weaker trend ($\beta\!=\!-0.019, p\!<\!0.05$, \textit{95\% c.i.}\ $-0.032\!:\!-0.005$), as measured with a linear model.
A possible explanation for the positive trend of constructions is related to production cost. Longer constructions are more costly for the speaker, so for them to still be an efficient production choice, their facilitating effect must be higher.

Finally, we observe that \textbf{referential constructions have a stronger facilitating effect than non-referential ones}.
Our LME model yields a positive effect for referentiality on \textit{FE} ($\beta\!=\!0.124, p\!<\!0.005$, 95\% c.i $0.099:0.149$) and we find a significant difference between the \textit{FE} of the two types ($t\!=\!3.115, p\!<\!0.005$, \textit{95\% c.i.}\ $0.072\!:\!0.016$).
Looking in more detail, first mentions of referential constructions have higher information content and lower \textit{FE} than first mentions of non-referential ones (\textit{H}: $t\!=\!15.435, p\!<\!0.005$, \textit{95\% c.i.}\ $1.115\!:\!0.864$; \textit{FE}: $t\!=\!-9.315, p\!<\!0.005$, \textit{95\% c.i.}\ $-0.246\!:\!-0.161$), perhaps since words in referential sequences tend to be less frequent and more context-dependent. However, when repeated, their information content drops more substantially, reproducing inverse frequency effects attested in humans for syntactic repetitions \cite{bock1986syntactic,scheepers2003syntactic}. As a result, their \textit{FE} exceeds that of non-referential constructions (\textit{FE}: $t\!=\!8.818, p\!<\!0.005$, \textit{95\% c.i.}\ $0.117\!:\!0.183$), with the information content of a repeated reference being 81\% lower than that of its utterance context.
Overall, these findings indicate that although referential constructions are less frequent than non-referential ones (23.3\% vs.\ 76.7\%; see §\ref{sec:extraction}), their repetition is a particularly effective strategy of information rate mitigation. \looseness=-1


\section{Discussion and Conclusions}
\label{sec:conclusion}

Construction repetition is a pervasive phenomenon in dialogue; their frequent occurrence gives constructions a processing advantage~\cite{conklin2012processing}.
In this paper, we show that the processing advantage of constructions can be naturally modelled as reduced information content and propose that speakers' production of constructions can be seen as a strategy for information rate mitigation. This strategy can explain why utterance information content is often found to decrease over the course of dialogues~\cite{vega2009looking,giulianelli2021analysing}, in contrast with the predictions of theories of optimal use of the communication channel~\cite{genzel2002entropy}.

We observe that, as predicted, construction usage in English open-domain spoken dialogues mitigates the information rate of utterances.
Furthermore, while constructions are produced at a stable rate throughout dialogues, their facilitating effect---our proposed measure of reduction in utterance information content---increases over time. We find that this increment is led by construction repetition, with facilitating effect being positively affected by repetition frequency, density, and by the contents of a construction. Repetitions of referential constructions reduce utterance information content more aggressively, arguably making them a more cost-reducing alternative to the shortening strategy observed in chains of referring expressions~\cite{krauss1964changes,krauss1967effect}, which instead tends to preserve rate constancy \cite{giulianelli2021dialogues}.\footnote{
	Expression shortening is more efficient, however, in terms of articulatory cost.
}\looseness-1

\paragraph{Relation to cognitive effort}
We consider repetitions as a way for speakers to make dialogic interaction less cognitively demanding both on the production and on the comprehension side. This is not at odds with the idea that repetitions are driven by interpersonal synergies \cite{fusaroli2014dialog} and coordination \cite{sinclair-fernandez-2021-construction}. We think that the operationalisation of these higher level processes can be described by means of lower level, efficiency-oriented mechanisms, with synergy and coordination both corresponding to reduced collaborative effort.
Although information content estimates from neural language models have been shown to correlate with human processing effort (cf.\ §\ref{sec:background-surprisal}), we cannot claim that our work directly models human cognitive processes as we lack the relevant human data to measure such correlation for the corpus at hand.

\paragraph{Adaptive language model}
Our decision to use an adaptive neural language model affects information content estimates in two main ways.
On the one hand, due to their high frequency, constructions are likely to be assigned higher probabilities by this model, and therefore lower information content. We stress that we do not present constructions’ lower information content as a novel result, nor do we make any claims based on this result.
As explained in §\ref{sec:preliminary-construction}, this is a precondition for our experiments on the facilitating effect of constructions, which is not determined exclusively by their information content (as empirically shown in §\ref{sec:ic-vs-fe}) but rather measures the effect of construction usage on the information content of entire utterances.
On the other hand, because our model is adaptive, the probability of constructions is likely to increase as a result of their appearance in the dialogue history. Adaptation, however, also contributes to lower utterance information content \textit{overall} through the exploitation of topical and stylistic cues, as demonstrated by the lower perplexity of the adaptive model on the entire target dialogue as well as on other dialogues from the same dataset (see §\ref{sec:modelling} and Appendix~\ref{sec:app-lr}).
In conclusion, while our adaptive language model assigns higher probabilities to frequently repeated tokens---as expected from a psychologically plausible model of utterance processing---it is not responsible for the discovered patterns of construction facilitating effect.
In future work, the model can be improved, e.g., by conditioning on the linguistic experience of individual speakers.

\paragraph{Types of dialogue} To consolidate our findings, construction repetition patterns should also be studied in dialogues of different genres and on datasets where utterance information content was not found to decrease.
We have chosen the Spoken BNC for our study as it contains dialogues from a large variety of real-life contexts, which makes it a representative dataset of open-domain dialogue.
In task-oriented dialogue, we expect constructions to consist of a more limited, task-specific vocabulary, resulting in longer chains of repetition and potentially more frequent referential construction usage.
These peculiarities of task-oriented dialogue may influence the strength of the facilitating effect (as we have seen, facilitating effect is affected by both frequency and referentiality) but we expect our main results to still hold, as they are generally related to the processing advantage of constructions. \looseness-1

\paragraph{Relevance for dialogue generation models}
Besides contributing new empirical evidence on construction usage in dialogue, our findings inform the development of more naturalistic utterance generation models.
They suggest that models should be continually updated for their probabilities to better reflect human expectations; that attention mechanisms targeting contexts of different sizes (local vs.\ global) may have a significant impact on the naturalness of generated utterances; and that while anomalous repetitions (e.g., generation loops) should be prevented \cite{li2016deep,holtzman2019curious}, it is important to ensure that natural sounding repetitions are not suppressed.
We expect dialogue systems that are able to produce human-like patterns of repetitions to be perceived as more natural overall---with users having the feeling that common ground is successfully maintained~\cite{PickeringGarrod2004}---and to lead to more effective communication~\cite{reitter2014alignment}.
In our view, such human-like patterns can be reproduced by steering generation models towards the trends of information rate observed in humans.


\section*{Acknowledgements}
We would like to thank the members of the Dialogue Modelling Group of the University of Amsterdam for their useful discussions, and our anonymous reviewers for their insightful comments. This project has received funding from the European Research Council (ERC) under the European Union's Horizon 2020 research and innovation programme (grant agreement No.\ 819455).

\bibliography{main}
\bibliographystyle{acl_natbib}

\appendix
\section*{Appendix}

\section{Possible Criteria to Distinguish Constructions}
\label{sec:app-criteria}
Lexicalised constructions can be classified according to multiple criteria \cite{titone1994descriptive,wray2002formulaic,columbus2013support}, including those listed below.
\begin{itemize}
    \item \textbf{Compositionality} This criterion is typically used to separate idioms from other formulaic expressions, although it is sometimes referred to as \textit{transparency} to underline its graded, rather than binary, nature. There is no evidence, however, that the processing advantage of idioms differs from that of compositional phrases \cite{tabossi2009idioms,jolsvai2013meaning,carrol2020all}. \textit{Therefore we ignore this criterion in the current study.}
    \item \textbf{Literal plausibility} This criterion is typically used to discriminate among different types of idioms \cite{titone1994descriptive,titone2014time}---as compositional phrases are literally plausible by definition. \textit{Because we ignore distinctions made on the basis of compositionality, we do not use this criterion.}
    \item \textbf{Meaningfulness} Meaningful expressions are idioms and compositional phrases (e.g. \textit{`on my mind'}, \textit{`had a dream'}) whereas sentence fragments that break constituency boundaries (e.g., \textit{`of a heavy'}, \textit{'by the postal'}) are considered less meaningful \cite[as measured in norming studies, e.g., by][]{jolsvai2013meaning}. There is some evidence that the meaningfulness of multi-word expressions correlates with their processing advantage even more than their frequency \cite{jolsvai2013meaning}; yet expressions are particularly frequent, they present processing advantages even if they break regular phrasal structures \cite{bybee1999effect,tremblay2011processing}. Moreover, utterances that break regular constituency rules are particularly frequent in spoken dialogue data (e.g., \textit{`if you could search for job and that's not'}, \textit{`you don't wanna damage your relationship with'}).
    \textit{For these reasons, we do not exclude constructions that span multiple constituents from our analysis.}
    \item \textbf{Schematicity} This criterion distinguishes expressions where all the lexical elements are fixed from expressions ``with slots'' that can be filled by varying lexical elements.\textit{In this study, we focus on fully lexicalised constructions.}
    \item \textbf{Familiarity} This is a subjective criterion that strongly correlates with objective frequency \cite{carrol2020all}. Human experiments would be required to obtain familiarity norms for our target data, and the resulting norms would only be an approximation of the familiarity judgements of the true speakers we analyse the language of. \textit{Therefore, we ignore this criterion in the current study.}
    \item \textbf{Communicative function} Formulaic expressions can fulfil a variety of discourse and communicative functions. \citet{biber2004if}, e.g., distinguish between stance expressions (attitude, certainty with respect to a proposition), discourse organisers (connecting prior and forthcoming discourse), and referential expressions; and for each of these three primary discourse functions, more specific subcategories are defined. This type of classification is typically done a posteriori---i.e., after a manual analysis of the expressions retrieved from a corpus according to other criteria \cite{biber2007lexical}. In the BNC, for example, we find epistemic lexical bundles (\textit{`I don't know'}, \textit{`I don't think'}), desire bundles (\textit{`do you want to'}, \textit{'I don't want to'}), obligation/directive bundles (\textit{`you don't have to'}), and intention/prediction bundles (\textit{`I'm going to'}, \textit{`it's gonna be'}). \textit{We do not use this criterion to avoid an a priori selection of the constructions.}
\end{itemize}

\section{Extraction of Repeated Constructions}
\label{sec:app-extraction}
\begin{table*}[!h]
\centering
\resizebox{\textwidth}{!}{
\begin{tabular}{cccclccc}
	\toprule
	\textbf{Speaker} & \textbf{RI} & \textbf{RI~Utt} & \textbf{Dist} & \textbf{Turn}  & $\boldsymbol{H(u)}$ & $\boldsymbol{H(c)}$ & $\boldsymbol{FE(c,u)}$ \\
	\midrule
A  &  0  &  0  &  -  &  [...]  I think that everyone should have the same opportunities & & & \\
					 & & & &	 \ \  and \textbf{I don't think you should be} proud or ashamed of what & 4.24 & 1.90 & 1.21 \\
					 & & & & 	\ \  your you know what your situation is whether you what your & & & \\
					 & & & &	   \ \   what your race is whether you're a woman or a man whether  & & & \\
					 & & & &	  \ \ you live from this pl whether you're in this place [...] & & & \\ \midrule
A & 1 & 0 & 80 &  I well I th I don't think it should \textbf{I don't think you should be} & 3.40 & 1.73 & 1.40 \\ \midrule
A  & 2 & 0 & 19    &  Well yes perhaps but \textbf{I don't think you should be} like um & 3.95 & 1.06 & 2.25 \\
			        & & & &	 \ \ embarrassed about it or I think I think you should just sort of &  & & \\ \bottomrule
\end{tabular}
}
\caption{Repetition chain for the construction \textit{`I don't think you should be'} in dialogue S2AX of the Spoken BNC, annotated with repetition index (RI), repetition index in utterance (RI~Utt), and distance from previous mention (Dist; number of tokens). $H(u)$ is the utterance information content, $H(c)$ and $FE(c,u)$ are the construction's information content and facilitating effect.}
\label{tab:chain-len7}
\end{table*}
We define a limited specific vocabulary of generic nouns that should not be considered referential. The vocabulary includes: \textit{bit}, \textit{bunch}, \textit{day}, \textit{days}, \textit{fact}, \textit{god}, \textit{idea}, \textit{ideas}, \textit{kind}, \textit{kinds}, \textit{loads}, \textit{lot}, \textit{lots}, \textit{middle}, \textit{ones}, \textit{part}, \textit{problem}, \textit{problems}, \textit{reason}, \textit{reasons}, \textit{rest}, \textit{side}, \textit{sort}, \textit{sorts}, \textit{stuff}, \textit{thanks}, \textit{thing}, \textit{things}, \textit{time}, \textit{times}, \textit{way}, \textit{ways}, \textit{week}, \textit{weeks}, \textit{year}, \textit{years}.
We also find all the filled pauses and exclude word sequences that consist for more than 50\% of filled pauses. Filled pauses in the Spoken BNC are transcribed as: \textit{huh}, \textit{uh}, \textit{erm}, \textit{hm}, \textit{mm}, \textit{er}.

\begin{figure}[h]
	\centering
	\includegraphics[width=\linewidth]{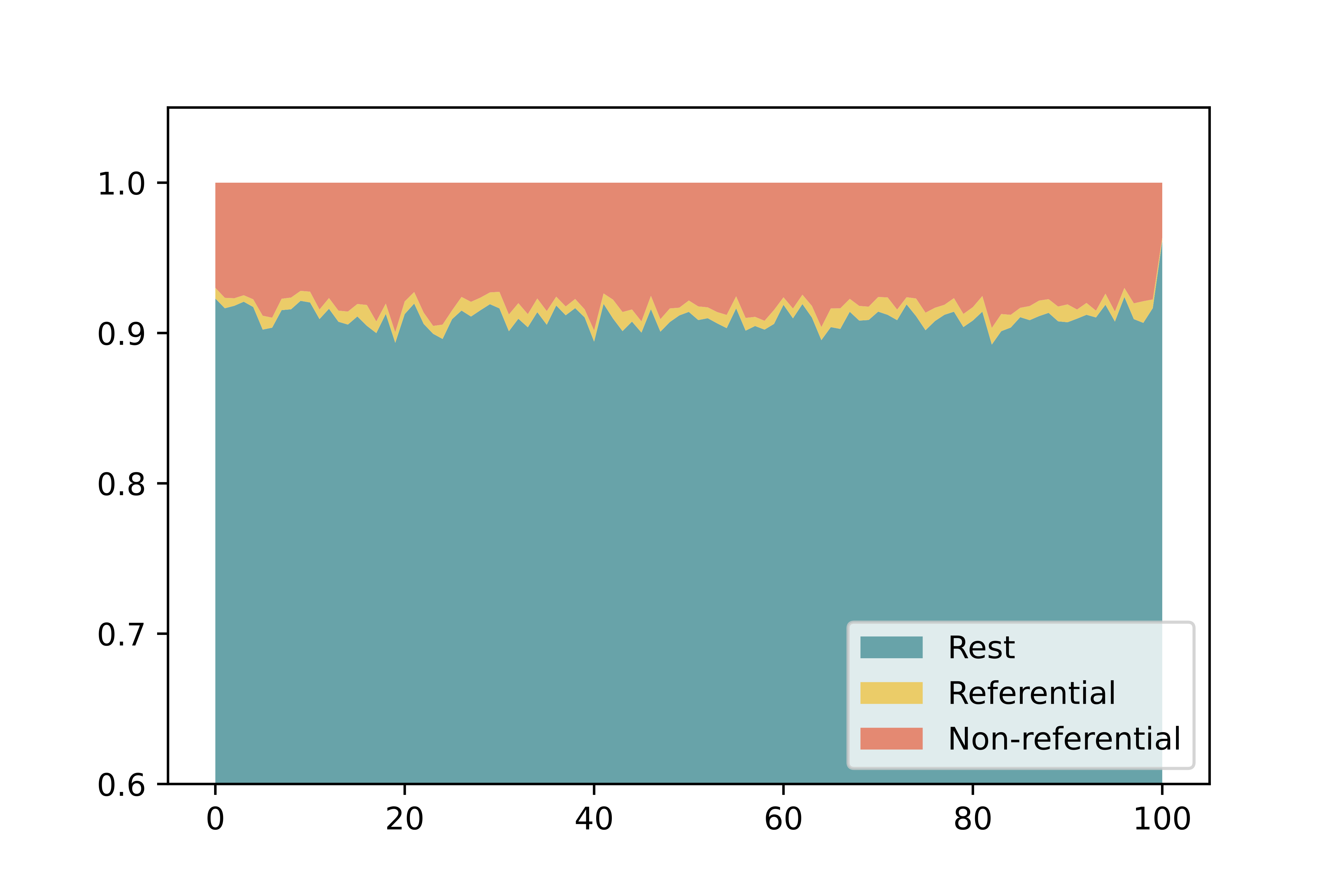}
	\caption{Proportion of tokens in an utterance that belong to referential constructions, non-referential constructions, and to non-construction sequences. The \textit{x} axis shows percentages indicating utterance positions in the dialogue relative to the dialogue length.}
	\label{fig:proportions}
\end{figure}

Figure~\ref{fig:proportions}  shows the proportion of tokens in an utterance belonging to constructions (referential and non-referential) and to non-construction sequences. Table~\ref{tab:chain-len7} shows a whole construction chain (from the first mention to the last repetition) for a construction of length 6.

\section{Language Model}
\subsection{Finetuning}
\label{sec:app-finetuning}
We finetune the \textit{`small' variant} of GPT-2 \cite{radford2019language} and DialoGPT \cite{zhang2019dialogpt} on our finetuning split of the Spoken BNC (see Section~\ref{sec:data}) using HuggingFace's implementation of the models with default tokenizers and parameters \cite{wolf2020transformers}. Dialogue turns are simply concatenated; we have experimented with labelling the dialogue turns (i.e., \textit{A: utterance 1, B: utterance 2} and found that this leads to higher perplexity. The finetuning results for both models are presented in Table~\ref{tab:finetuning}.
We finetune the models and measure their perplexity using Huggingface's finetuning script. We use early stopping over 5 epochs.\footnote{The number of epochs (5) has been selected in preliminary experiments together with the learning rate (\num{1e-4}). In these experiments---which we ran for 40 epochs---we noticed that the \num{1e-4} learning rate offers the best tradeoff of training time and perplexity out of four possible values: \num{1e-2}, \num{1e-3}, \num{1e-4}, \num{1e-5}. We obtained insignificantly lower perplexity values with a learning rate of \num{1e-5}, with significantly longer training time: 20 epochs for GPT-2 and 28 epochs for DialoGPT.} Sequence length and batch size vary together because they together determine the amount of memory required; more expensive combinations (e.g., 256 tokens with batch size 16) require an exceedingly high amount of GPU memory. Reducing the maximum sequence length has limited impact: 99.90\% of dialogue turns have at most 128 words.

DialoGPT starts from extremely high perplexity values but catches up quickly with finetuning. GPT-2 starts from much lower perplexity values and reaches virtually the same perplexity as DialoGPT after finetuning.
For the pre-trained DialoGPT perplexity is extremely high, and the perplexity trend against maximum sequence length is surprisingly upward.
These two behaviours indicate that the pre-trained DialoGPT is less accustomed than GPT-2 to the characteristics of our dialogue data. DialoGPT is trained on written online group conversations, while we use a corpus of transcribed spoken conversations between two speakers. In contrast, GPT-2 has been exposed to the genre of fiction, which contains scripted dialogues, and thus to a sufficiently similar language use. We select GPT-2 finetuned with a maximum sequence length of 128 and 512 as our best two models; these two models (which we now refer to as \textit{frozen}) are used for the adaptive learning rate selection (Section~\ref{sec:app-lr}).

\begin{table*}[]
\centering
\resizebox{\textwidth}{!}{
\begin{tabular}{lcccccc}
\toprule
\textbf{Model} & \textbf{Learning rate} & \textbf{Max sequence length} & \textbf{Batch size} & \textbf{Best epoch} & \textbf{Perplexity finetuned} & \textbf{Perplexity pre-trained} \\ \midrule
DialoGPT       & 0.0001                 & 128                          & 16                  & 3                   & 23.21                        & 7091.38                       \\
DialoGPT       & 0.0001                 & 256                          & 8                   & 4                   & 22.26                        & 12886.92                      \\
DialoGPT       & 0.0001                 & 512                          & 4                   & 4                   & 21.73                        & 21408.32                      \\
GPT-2          & 0.0001                 & 128                          & 16                  & 4                   & 23.32                        & 173.76                        \\
GPT-2          & 0.0001                 & 256                          & 8                   & 3                   & 22.21                        & 159.23                        \\
GPT-2          & 0.0001                 & 512                          & 4                   & 3                   & 21.55                        & 149.82               \\ \bottomrule
\end{tabular}
}
\caption{Finetuning results for GPT-2 and DialoGPT on our finetuning split of the Spoken BNC.}
\label{tab:finetuning}
\end{table*}

\subsection{Learning Rate Selection}
\label{sec:app-lr}
To find the appropriate learning rate for on-the-fly adaptation (see Section~\ref{sec:modelling}), we randomly select 18 dialogues $D$ from the analysis split of the Spoken BNC and run an 18-fold cross-validation for a set of six candidate learning rates: \num{1e-5}, \num{1e-4}, $\ldots$, 1. We finetune the model on each dialogue using one of these learning rate values, and compute perplexity change 1) on the dialogue itself (to measure \textit{adaptation}) as well as 2) on the remaining 17 dialogues (to measure \textit{generalisation}). We set the Transformer's context window to 50 to reproduce the experimental conditions presented in Section~\ref{sec:estimates}.

More precisely, for each dialogue $d \in D$, we calculate the perplexity of our two frozen models (Section~\ref{sec:app-finetuning}) on $d$ and $D \setminus \{d\}$ (which we refer to as $ppl_{before}(d)$ and $ppl_{before}(D)$, respectively). Then, we finetune the models on $d$ using the six candidate learning rates, and measure again the perplexity over $d$ and $D \setminus \{d\}$ (respectively, $ppl_{after}(d)$ and $ppl_{after}(D)$). The change in performance is evaluated according to two metrics: $\frac{ppl_{after}(d) - ppl_{before}(d)}{ppl_{before}(d)}$ measures the degree to which the model has successfully adapted to the target dialogue; $\frac{ppl_{after}(D) - ppl_{before}(D)}{ppl_{before}(D)}$ measures whether finetuning on the target dialogue has caused any loss of generalisation.

\begin{figure}[h]
\centering
\begin{subfigure}{.23\textwidth}
  \centering
  \includegraphics[width=\linewidth]{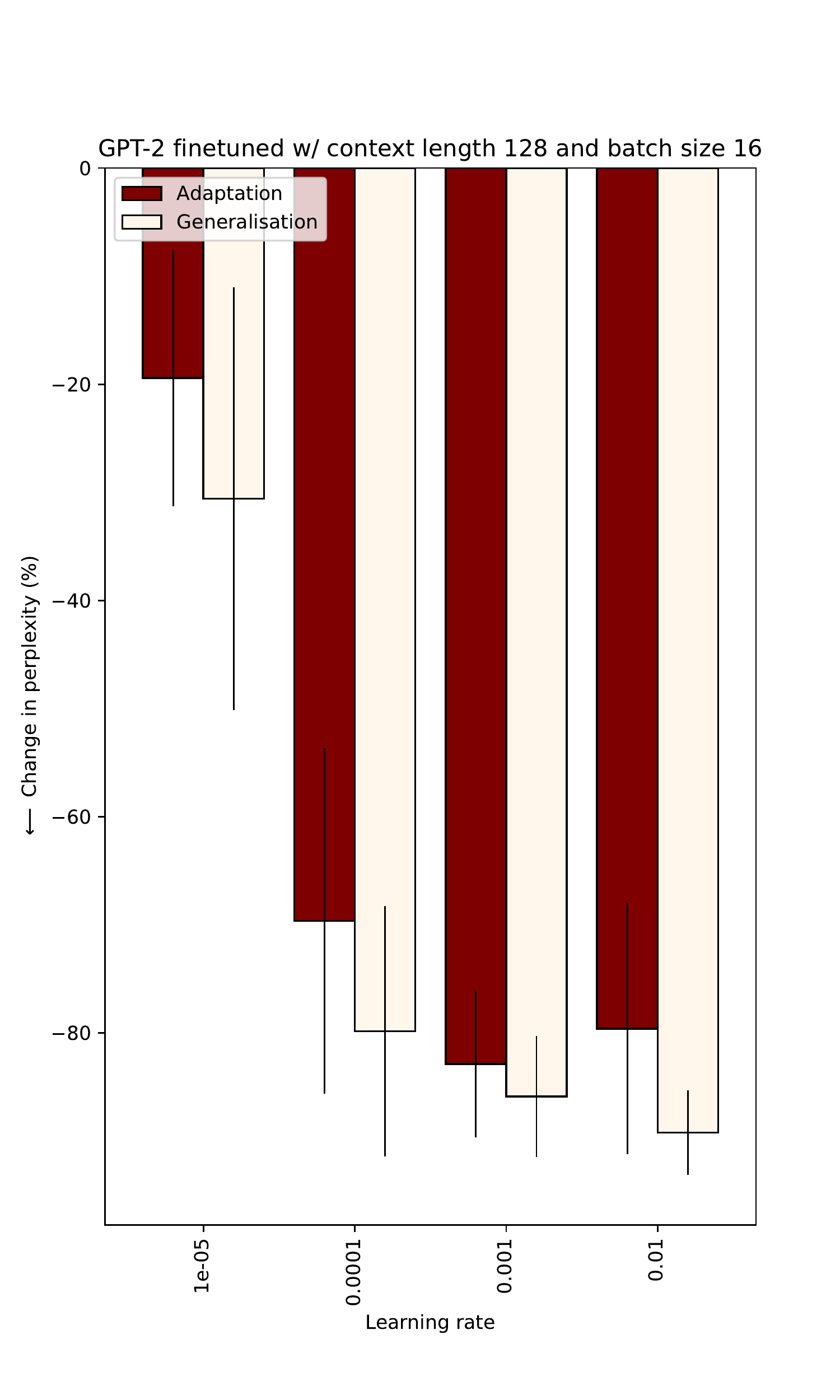}
  \vspace{-2.5em}
  \caption{}
  \label{fig:lr-128}
\end{subfigure}%
\hspace{.01\textwidth}
\begin{subfigure}{.23\textwidth}
  \centering
  \includegraphics[width=\linewidth]{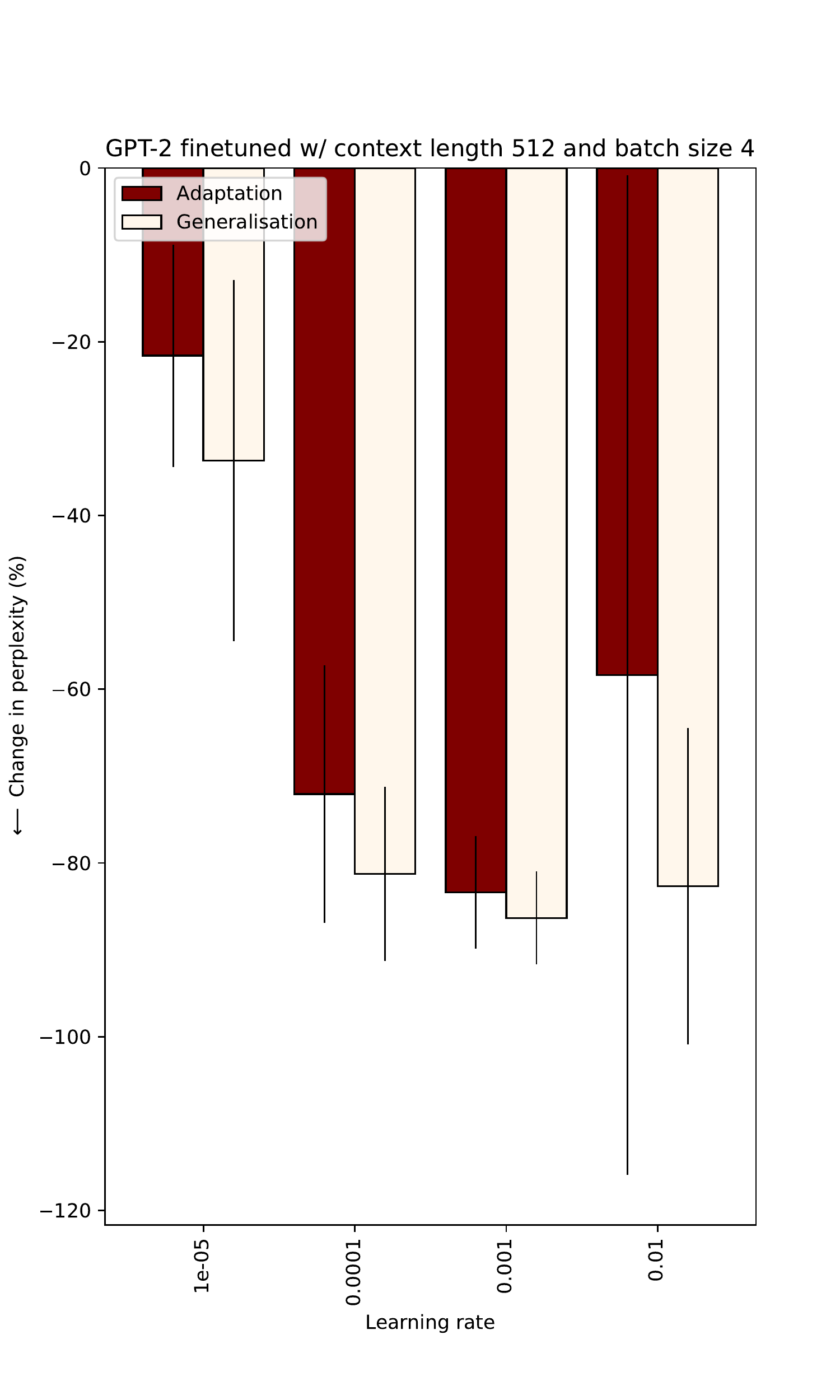}
  \vspace{-2.5em}
  \caption{}
  \label{fig:lr-512}
\end{subfigure}%
\caption{The adaptation and generalisation performance (defined in Section~\ref{sec:app-lr}) with varying learning rate.}
\label{fig:lr}
\end{figure}

The learning rate selection results are presented in Figure~\ref{fig:lr}. We select \num{1e-3} as the best learning rate and pick the model finetuned with a maximum sequence length of 512 as our best model. The difference in perplexity reduction (both adaptation and generalisation) is minimal with respect to the model finetuned with a maximum sequence length of 128, but since the analysis split of the Spoken BNC contains turns longer than 128 tokens, we select the 512 version.
Similarly to \citet{van2018neural}, we find that finetuning on a dialogue does not cause a loss in generalisation but instead helps the model generalise to other dialogues. Unlike \shortcite{van2018neural}, who used LSTM language models, we find that learning rates larger than $\num{1e-1}$ cause backpropagation to overshoot, even within a single dialogue. In Figure~\ref{fig:lr}, the bars for $\num{1e-1}$ and $1$ are not plotted because the corresponding data contains infinite perplexity values (due to numerical overflow). The selected learning rate, $\num{1e-3}$, is a relatively low learning rate for on-the-fly adaptation but it is still higher than the best learning rate for the entire dataset by a factor of 10.

\section{Linear Mixed Effect Models}
\label{sec:app-lme}
\begin{figure}[h]
	\centering
	\includegraphics[width=0.9\linewidth]{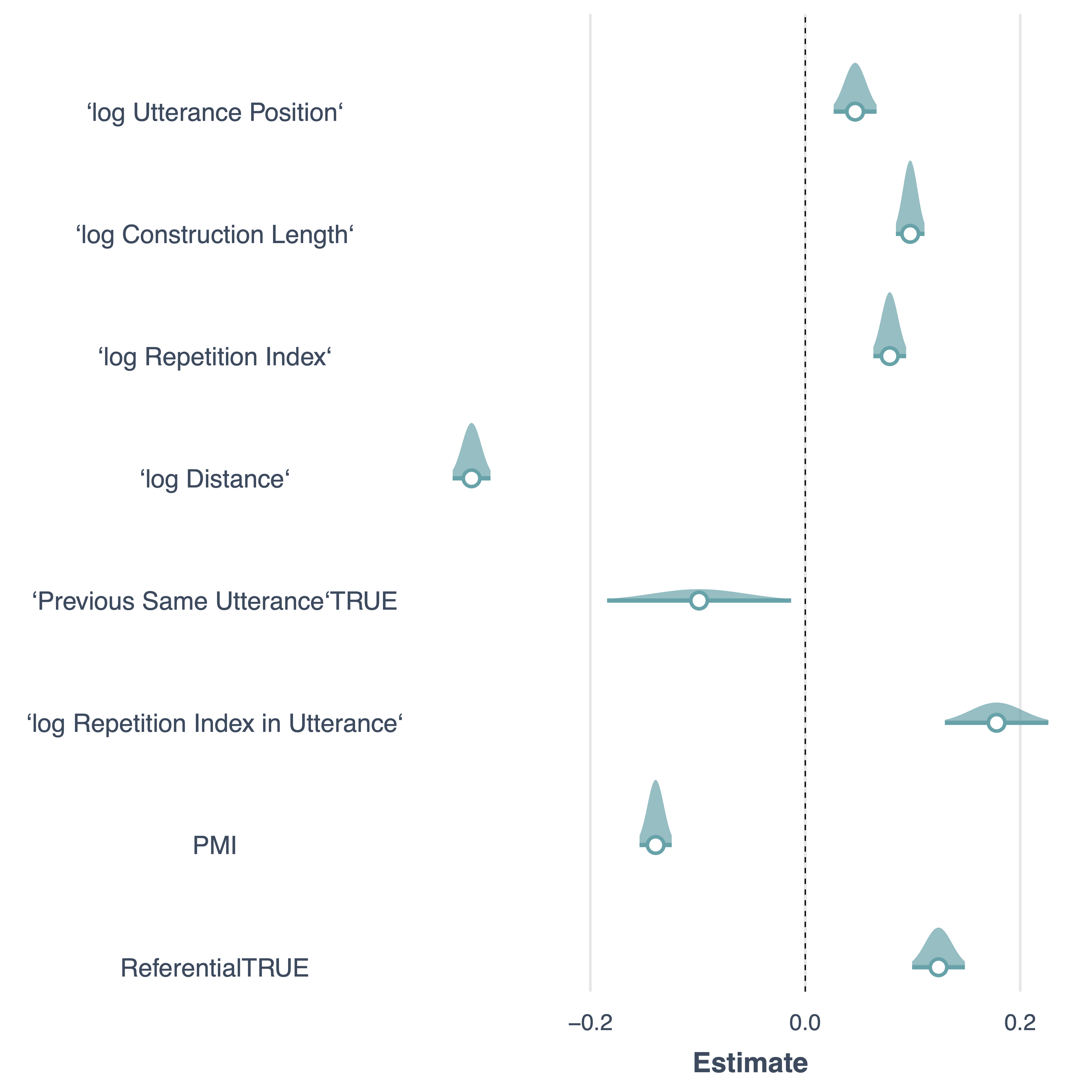}
	\caption{Significant predictors of facilitating effect. Mixed effects linear regression, continuous predictors are mean-centred and scaled by 2 standard deviations.}
	\label{fig:fe-lme-summary}
\end{figure}

As explained in §\ref{sec:stat-model} of the main paper, we fit a linear mixed effect model using facilitating effect as the response variable and including multilevel random effects grouped by dialogues and individual speakers.\footnote{We also try grouping observations only by dialogue and only by individual speakers. The amount of variance explained (but unaccounted for by the fixed effects) decreases, so we keep the two-level random effects.}. The fixed effects of the model, resulting from a backward stepwise selection procedure, are presented in §\ref{sec:stat-model}. Non-binary predictors are log-transformed, mean-centered, and scaled by 2~sd. The final model is summarised in Listing~\ref{lme-fe} and its coefficients are visualised in Figure~\ref{fig:fe-lme-summary}.
We rely on the \texttt{lme4} and \texttt{lmerTest} R packages for this analysis.

\begin{lstlisting}[label=lme-fe,float=*,frame=tb,caption=Linear mixed effect model for Facilitating Effect]

MODEL INFO:
Observations: 46399
Dependent Variable: Facilitating Effect
Type: Mixed effects linear regression

MODEL FIT:
AIC = 99197.283, BIC = 99302.224
Pseudo-R^2 (fixed effects) = 0.084
Pseudo-R^2 (total) = 0.111

FIXED EFFECTS:
-----------------------------------------------------------------------------------
                               Est.     2.5%    97.5%    t val.        d.f.       p
--------------------------- ------- -------- -------- --------- ----------- -------
(Intercept)                   0.704    0.683    0.725    65.527     185.698   0.000
log Utterance Position        0.046    0.026    0.066     4.556    9274.269   0.000
log Construction Length       0.098    0.084    0.111    14.396   46372.022   0.000
log Repetition Index          0.079    0.063    0.094    10.096   45082.205   0.000
log Distance                 -0.311   -0.328   -0.293   -34.571   46269.156   0.000
Previous Same Utterance      -0.099   -0.184   -0.013    -2.262   46063.723   0.024
log Rep. Index in Utterance   0.178    0.130    0.226     7.243   45765.367   0.000
PMI                          -0.139   -0.154   -0.124   -18.225   45172.205   0.000
Referential                   0.124    0.099    0.149     9.887   46214.616   0.000
-----------------------------------------------------------------------------------

p values calculated using Satterthwaite d.f.

RANDOM EFFECTS:
------------------------------------------------
Group                    Parameter    Std. Dev.
---------------------- ------------- -----------
Speaker:`Dialogue ID    (Intercept)     0.082
     Dialogue ID        (Intercept)     0.090
       Residual                         0.701
------------------------------------------------

Grouping variables:
-----------------------------------------
Group                   # groups    ICC
---------------------- ---------- -------
Speaker:`Dialogue ID      368      0.013
    Dialogue ID           185      0.016
-----------------------------------------

Continuous predictors are mean-centered and scaled by 2 s.d.
\end{lstlisting}

\section{Further Results}
\label{sec:app-further-results}

\subsection{Same-Utterance Self-Repetitions}
\label{sec:app-same-utterance}
We investigate the interaction between cumulativity and recency (see §\ref{sec:results-mentions}) by focusing on densely clustered repetitions, produced by a speaker within a single utterance (the median distance between repetitions in the same utterance is 8 words; across turns it is 370.5 words). Table~\ref{tab:chain-len3} shows an example of same-utterance repetition.
Repeating a construction when it has already been mentioned in the current utterance limits its facilitating effect ($\beta=-0.099, p<0.05$, \textit{95\% c.i.} -0.184:-0.013): if a portion of the utterance already consists of a construction, utterance information content will already be reduced, which in turn reduces the potential for the facilitating effect of repetitions.
Nevertheless, we find \textbf{strong cumulativity effects for self-repetitions within the same utterance}: the repetition index \textit{within the current utterance} of a construction mention (i.e., how often the construction has been repeated so far in the utterance) has a positive effect on \textit{FE} ($\beta=0.178, p<0.005$, \textit{95\% c.i.} 0.130:0.226); see Figure~\ref{fig:fe-inturn}.
In sum, same-utterance self-repetitions, especially those involving three or more mentions in a single utterance, can have a strong reduction effect on utterance information content. Although this may seem a simple yet very effective strategy for information rate mitigation, it is unlikely to be very effective in terms of the amount of information exchanged. Indeed, speakers do not use this strategy often in the Spoken BNC: 6.82\% of the total construction occurrences have at least one previous mention in the same utterance.

\subsection{Interaction-Specificity}
\begin{figure}[h]
	\centering
	\begin{subfigure}{.23\textwidth}
		\centering
		\includegraphics[width=\linewidth]{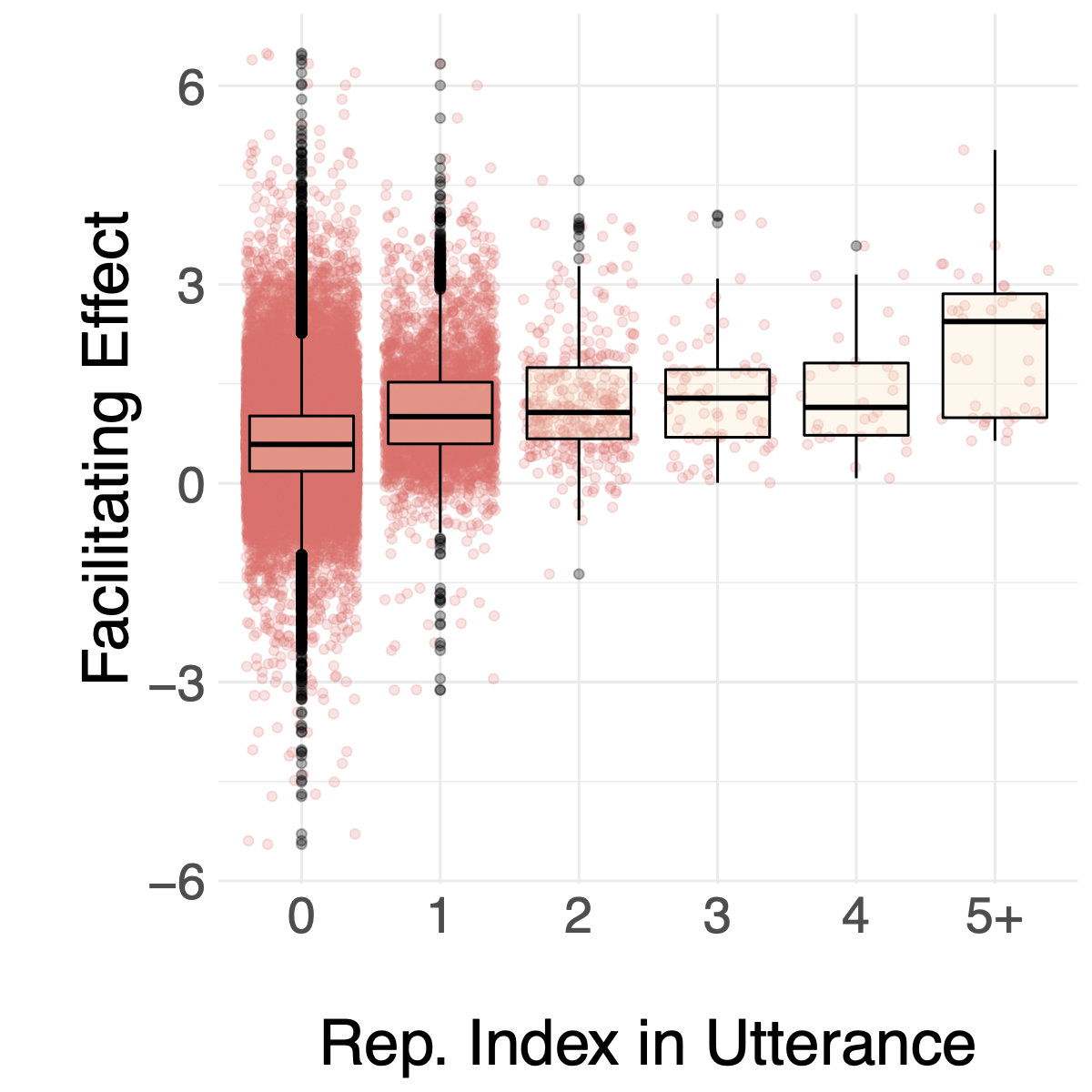}
		\caption{}
		\label{fig:fe-inturn}
	\end{subfigure}
	\begin{subfigure}{.23\textwidth}
		\centering
		\includegraphics[width=\linewidth]{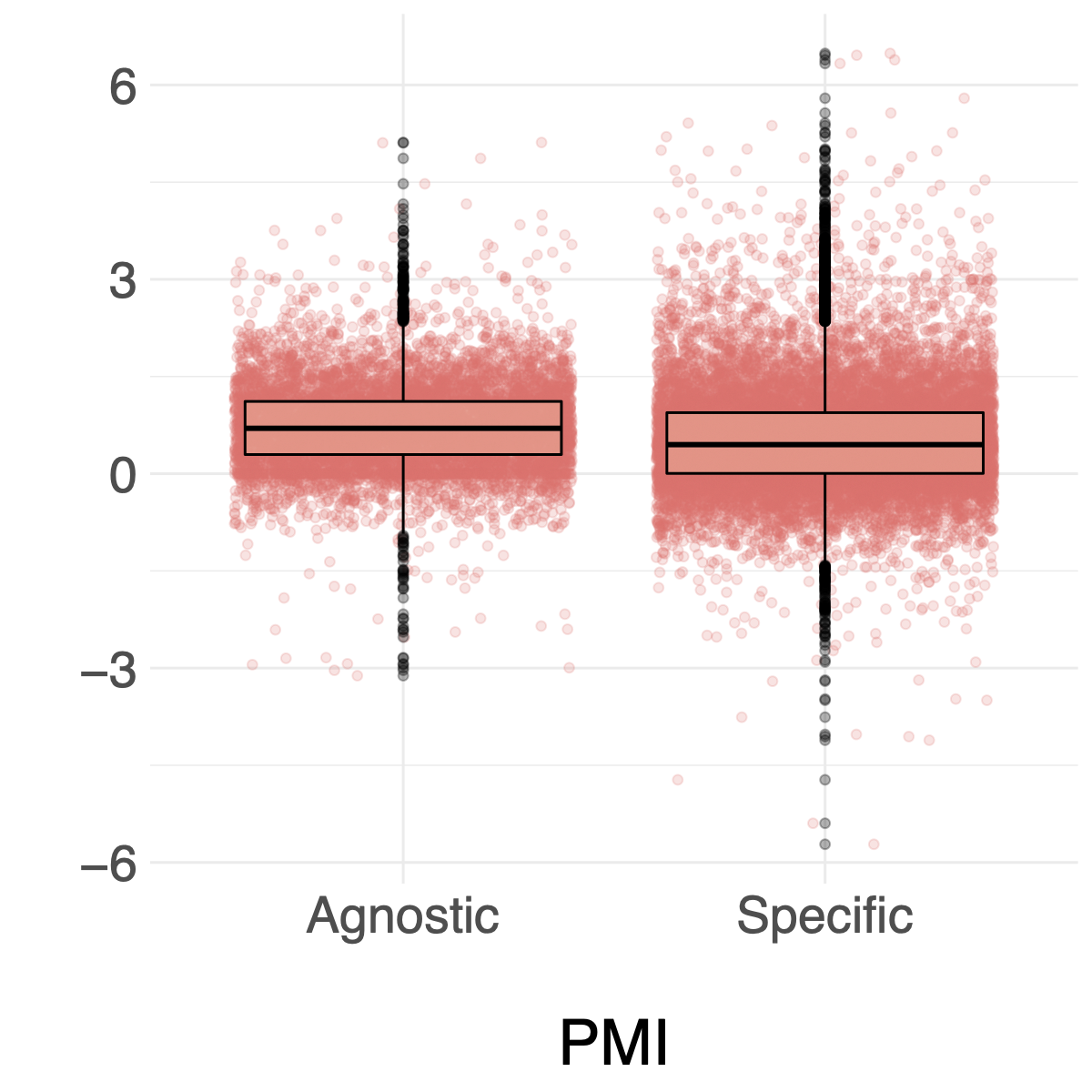}
		\caption{}
		\label{fig:fe-pmi}
	\end{subfigure}
	\caption{Facilitating effect against repetition index \textit{within the current utterance} (a) and facilitating effect of interaction-agnostic constructions ($\operatorname{PMI}(c,d)< 1$) vs.\ interaction-specific constructions ($\operatorname{PMI}(c,d) = \max_{c',d'} \operatorname{PMI}(c', d')$) (b).}
\end{figure}

To distinguish interaction-specific constructions---those repeated particularly often in certain dialogues---from interaction-agnostic ones,
we measure the association strength between a construction $c$ and a dialogue $d$ as the pointwise mutual information (PMI) between the two:
\begin{align}
	\operatorname{PMI}(c,d) = \log_2 \frac{P(c|d)}{P(c)}
	\label{eq:pmi}
\end{align}
This quantifies how unusually frequent a construction is in a given dialogue, compared to the rest of the corpus. For example, for a construction to obtain a PMI score of 1, its probability given the dialogue $P(c|d)$ must be twice as high as its prior probability $P(c)$. Low PMI scores (especially below 1) characterise interaction-agnostic constructions, whereas higher PMI scores indicate that constructions are specific to a given dialogue.
The probabilities in Eq.~\ref{eq:pmi} are obtained using maximum likelihood estimation over the analysis split of the Spoken BNC.
PMI scores have a negative effect on \textit{FE} ($\beta=-0.139, p<0.005$, 95\% c.i. -0.154:-0.124), indicating that interaction-agnostic constructions have a stronger facilitating effect than interaction-specific ones.
Figure~\ref{fig:fe-pmi} shows the \textit{FE} distributions for the most extreme cases: constructions with a PMI lower than 1 (`agnostic') and constructions that have been repeated in only one dialogue (`specific').

\section{Computing Infrastructure and Budget}
\label{sec:app-compute}
Our experiments were carried out using a single GPU on a computer cluster with Debian Linux OS. The GPU nodes on the cluster are GPU GeForce 1001 1080Ti, 11GB GDDR5X, with NVIDIA driver version 418.56 and CUDA version 10.1. The total computational budget required to finetune the language model amounts to 45 minutes; obtaining surprisal estimates requires 4 hours, and selecting the adaptation learning rate requires 9 hours.

\end{document}